\documentclass[letterpaper]{article} % DO NOT CHANGE THIS
\usepackage{aaai2026}  % DO NOT CHANGE THIS
\nocopyright
\usepackage{times}  % DO NOT CHANGE THIS
\usepackage{helvet}  % DO NOT CHANGE THIS
\usepackage{courier}  % DO NOT CHANGE THIS
\pdfoutput=1
\usepackage[hyphens]{url}  % DO NOT CHANGE THIS
\usepackage{graphicx} % DO NOT CHANGE THIS
\usepackage{natbib}  % DO NOT CHANGE THIS AND DO NOT ADD ANY OPTIONS TO IT
\usepackage{caption} % DO NOT CHANGE THIS AND DO NOT ADD ANY OPTIONS TO IT
\usepackage{algorithm}

\usepackage[utf8]{inputenc} % allow utf-8 input
\usepackage{booktabs}       % professional-quality tables
\usepackage{nicefrac}       % compact symbols for 1/2, etc.
\usepackage{microtype}      % microtypography
\usepackage{makecell}
\usepackage[table]{xcolor}
\definecolor{myblue}{RGB}{206,230,247}

\usepackage{amssymb}
\usepackage{mathtools}
\usepackage{amsthm}
\usepackage{multirow}

\theoremstyle{plain}

\theoremstyle{definition}

\theoremstyle{remark}

\usepackage{algpseudocode}

\usepackage{subcaption}

\usepackage{pifont}

\newcommand{\cmark}{\ding{51}}  % ✓
\newcommand{\xmark}{\ding{55}}  % ✗

\usepackage{paralist}
\usepackage{newfloat}
\usepackage{listings}
\DeclareCaptionStyle{ruled}{labelfont=normalfont,labelsep=colon,strut=off} % DO NOT CHANGE THIS
\floatstyle{ruled}
\newfloat{listing}{tb}{lst}{}
\floatname{listing}{Listing}

\title{FARM: Frame-Accelerated Augmentation and Residual Mixture-of-Experts for Physics-Based High-Dynamic Humanoid Control}
\author {
    Tan Jing\textsuperscript{\rm 1},
    Shiting Chen\textsuperscript{\rm 2},
    Yangfan Li\textsuperscript{\rm 1},
    Weisheng Xu\textsuperscript{\rm 1},
    Renjing Xu\textsuperscript{\rm 1}
}
\affiliations {
    \textsuperscript{\rm 1} Hong Kong University of Science and Technology (Guangzhou), China \\
    \textsuperscript{\rm 2} Guangdong University of Technology, China \\
    Emails: jingt@hkust-gz.edu.cn \\
    Corresponding author: renjingxu@hkust-gz.edu.cn
}

\begin{document}

\maketitle

\begin{abstract}
% Unified physics-based humanoid controllers are pivotal for robotics and character animation, yet models that excel on gentle, everyday motions still stumble on explosive actions, hampering real-world deployment. 
% We bridge this gap with High-Dynamic Humanoid Control (HDHC), an end-to-end framework composed of frame-accelerated augmentation, a robust base controller, and a residual mixture-of-experts (MoE). Frame-accelerated augmentation exposes the model to high-velocity pose changes by widening inter-frame gaps. The base controller reliably tracks everyday low-dynamic motions, while the residual MoE adaptively allocates additional network capacity to handle challenging high-dynamic actions, significantly enhancing tracking accuracy.
% In the absence of a public benchmark, we curate the High-Dynamic Humanoid Motion (HDHM) dataset, comprising 3593 physically plausible clips. 
% On HDHM, HDHC reduces the tracking failure rate by 42.8\% and lowers global mean per-joint position error by 14.6\% relative to the baseline, while preserving near-perfect accuracy on low-dynamic motions. 
% These results establish HDHC as a new baseline for high-dynamic humanoid control and introduce the first open benchmark dedicated to this challenge.

Unified physics-based humanoid controllers are pivotal for robotics and character animation, yet models that excel on gentle, everyday motions still stumble on explosive actions, hampering real-world deployment.
We bridge this gap with FARM (Frame-Accelerated Augmentation and Residual Mixture-of-Experts), an end-to-end framework composed of frame-accelerated augmentation, a robust base controller, and a residual mixture-of-experts (MoE). Frame-accelerated augmentation exposes the model to high-velocity pose changes by widening inter-frame gaps. The base controller reliably tracks everyday low-dynamic motions, while the residual MoE adaptively allocates additional network capacity to handle challenging high-dynamic actions, significantly enhancing tracking accuracy.
In the absence of a public benchmark, we curate the High-Dynamic Humanoid Motion (HDHM) dataset, comprising 3593 physically plausible clips.
On HDHM, FARM reduces the tracking failure rate by 42.8\% and lowers global mean per-joint position error by 14.6\% relative to the baseline, while preserving near-perfect accuracy on low-dynamic motions.
These results establish FARM as a new baseline for high-dynamic humanoid control and introduce the first open benchmark dedicated to this challenge. The code and dataset will be released at https://github.com/Colin-Jing/FARM.
\end{abstract}

\section{Introduction}\label{sec:intro}
% 大背景大领域，引出大背景大领域的现存问题
Physics-based humanoid motion imitation combines the visual fidelity of motion-capture data with the strict dynamics of rigid-body simulation, underpinning advanced robotics, digital characters, and immersive AR/VR experiences. In practice, however, each new target sequence still demands days of reward redesign and hyperparameter tuning. This heavy engineering overhead hampers large-scale adoption and highlights the need for a robust universal controller that can track a wide range of motions without per-task retraining.

The Universal Humanoid Controller (UHC) \cite{luo2021dynamics} first explored the challenge of universal motion tracking, achieving a 97\% tracking success rate on the 10k-sequence AMASS \cite{AMASS} dataset by incorporating a hand-crafted residual force controller (RFC) \cite{yuan2020residual}. Perpetual Humanoid Control (PHC) \cite{luo2023perpetual} advanced this by removing the RFC and introducing a progressive multiplicative control architecture, elevating the tracking success rate to 98.9\%. Its subsequent refinement, PHC+ \cite{luo2024universal}, further improved performance through data filtering and training optimizations, achieving near-100\% success. However, the incremental learning approach inherent in PHC+ complicates the training process and hampers scalability. Addressing this complexity, the Fully Constrained (FC) Controller proposed in MaskedMimic \cite{tessler2024maskedmimic} adopts a unified network architecture, simplifying training and yielding superior generalization capabilities. Despite these methodological advances, all aforementioned controllers share a critical limitation in that they are trained primarily on the AMASS dataset, dominated by simple, low-dynamic everyday motions. This restricts their ability to generalize to highly dynamic motions, leaving the robust control of explosive actions an open challenge in universal humanoid control.

Building upon this limitation, we initially considered a straightforward intuition: since the AMASS dataset predominantly comprises low-dynamic actions with small pose-to-pose variations, artificially accelerating motion clips through uniform frame dropping might enlarge the pose differences between successive frames, mimicking the abrupt transitions seen in highly dynamic motions. However, experiments (Section \ref{subsec:ablation}) revealed that fine-tuning the FC controller directly on these frame-accelerated clips provided minimal improvement for high-dynamic actions, and even degraded its previously near-perfect accuracy on low-dynamic sequences. This prompted a second key insight inspired by human motor attention: routine, everyday movements demand minimal cognitive resources, whereas explosive, high-dynamic actions require heightened attention and additional processing capacity. Analogously, we reasoned that effective high-dynamic control necessitates dynamically adjusting network capacity to match motion intensity, ensuring high accuracy without compromising low-dynamic performance.

\begin{figure*}[ht]
  \centering
  \includegraphics[width=\linewidth]{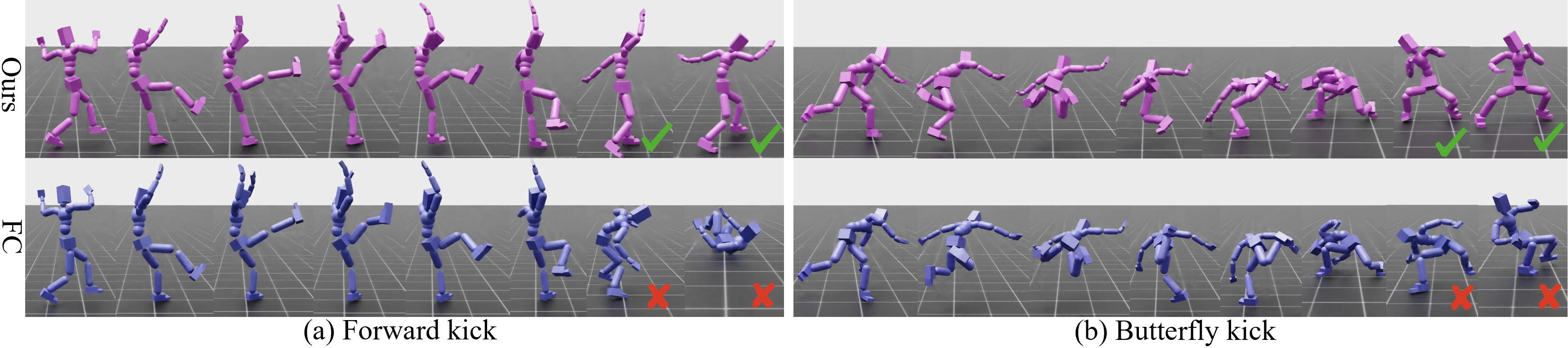}
  \caption{Comparison between FARM and the baseline FC on two high-dynamic motions. FARM accurately completes both motions (green check), while FC falls or loses balance (red cross). Frames are shown in playback order from left to right.}
  %: forward kick and butterfly kick
  % \caption{Motion frames are presented in playback order from left to right. FARM accurately tracks four distinct high‑dynamic motions not encountered during training, demonstrating its superior generalization capability.}
  \label{fig:intro_overlay}
\end{figure*}

Motivated by these insights, we propose the FARM framework, which consists of three components: frame-accelerated augmentation, a base controller, and a residual mixture-of-experts (MoE). Frame-accelerated augmentation widens the pose gap between successive frames to expose the model to the abrupt transitions found in highly dynamic actions. The base controller delivers dependable baseline motion tracking for everyday movements. Building on this foundation, the residual MoE tackles the remaining high-dynamic errors. Its speed-aware router (SAR) stratifies motions by dynamic intensity and directs each band to a dedicated subset of experts, while the dynamic expert-assignment (DEA) adaptively scales capacity, activating only the required experts for low-dynamic segments and progressively more for highly dynamic ones. Together, these components enable FARM to master challenging high-dynamic motions while preserving strong performance on low-dynamic tasks.
Figure~\ref{fig:intro_overlay} illustrates FARM’s ability to accurately complete complex high-dynamic motions, where the baseline FC model fails.

% Motivated by these insights, we propose the High-Dynamic Humanoid Control (FARM) framework, which consists of three components: frame-accelerated augmentation, a base controller, and a residual mixture-of-experts (MoE). Frame-accelerated augmentation widens the pose gap between successive frames to expose the model to the abrupt transitions found in highly dynamic actions. The base controller delivers dependable baseline motion tracking for everyday movements. Building on this foundation, the residual MoE tackles the remaining high-dynamic errors. Its speed-aware router (SAR) stratifies motions by dynamic intensity and directs each band to a dedicated subset of experts, while the dynamic expert-assignment (DEA) adaptively scales capacity, activating only the required experts for low-dynamic segments and progressively more for highly dynamic ones. Together, these components enable FARM to master challenging high-dynamic motions while preserving strong performance on low-dynamic tasks. Figure  \ref{fig:intro_overlay} shows that FARM accurately tracks unseen high-dynamic motions.

Because no public dataset exists specifically for high-dynamic humanoid motions, we curate the High-Dynamic Humanoid Motion (HDHM) benchmark for controlled evaluation. HDHM comprises 3593 physics-plausible clips, each manually screened to ensure kinematic validity. On this benchmark, our FARM framework cuts the tracking failure rate by 42.8\% and lowers the global mean per-joint position error (MPJPE$_g$) by 14.6\% relative to the baseline, while preserving the near-perfect accuracy of the baseline on low-dynamic motions.

\begin{enumerate}
  \item \textbf{HDHM dataset.} We release the first open, manually curated benchmark dedicated to high-dynamic humanoid control, comprising 3593 physics-plausible clips.
  
  \item \textbf{FARM framework.} We propose a frame-accelerated residual MoE framework. Frame-accelerated augmentation allows the model to acquire high-dynamic skills from predominantly low-dynamic data, and the residual MoE adaptively allocates computational capacity according to motion intensity.
  
  \item \textbf{Comprehensive experiments.} Extensive evaluations show that FARM significantly outperforms strong baselines on challenging high-dynamic motions while preserving the near-perfect accuracy on low-dynamic motions.
\end{enumerate}

\section{Related Works}\label{sec:related}

\subsection{Physics-based motion imitation}
Physics-based motion imitation learns control policies to track mocap references in simulation using reinforcement learning \cite{mnih2016asynchronous}. DeepMimic~\cite{peng2018deepmimic}, AMP~\cite{peng2021amp}, and PHC~\cite{luo2023perpetual} progressively improved motion diversity and generalization, with PHC+~\cite{luo2024universal} achieving near-perfect tracking on large-scale datasets like AMASS. MaskedMimic~\cite{tessler2024maskedmimic} further unified diverse conditioning forms into a single controller via masked modeling. While effective on standard mocap data, these methods struggle with highly dynamic motions due to limited exposure to fast transitions. Our approach addresses this by explicitly targeting high-dynamic regimes through data and architectural adaptations.

\subsection{Mixture-of-experts in motion control}
Several recent works apply MoE to improve specialization and scalability in humanoid or legged control. GMT~\cite{chen2025gmt} employs MoE alongside adaptive sampling to track diverse whole-body motions with a unified policy. MoRE~\cite{wang2025more} uses a mixture of latent residual experts to learn human-like gaits across complex terrains. MoE‑Loco~\cite{huang2025moe} addresses multitask locomotion with dynamic selection of experts to mitigate gradient conflicts and support varied gait and terrain combinations. 
These methods demonstrate the benefits of MoE-based specialization. Unlike fixed expert usage in prior work, our residual MoE adapts the expert count based on motion intensity for better capacity allocation.

\section{Method}\label{sec:method}

% This section details how we convert the Full-Controller (FC) backbone into High-Dynamic Humanoid Control (FARM).  We first give an overview of the complete framework and its training flow, then recap the original FC architecture, introduce the frame-accelerated data augmentation that exposes the controller to high-energy trajectories, and finally describe the residual mixture-of-experts that corrects FC in distinct speed bands while keeping inference latency unchanged.

% More concretely, Section 3.1 outlines the overall pipeline shown in Figure 2; Section 3.2 revisits the FC design so that later modifications are self-contained; Section 3.3 explains how random acceleration and uniform subsampling create failure-focused training data; Section 3.4 presents the residual experts that learn speed-specific embedding corrections, followed by the gating mechanism that selects one expert per step and runs it in parallel with the frozen FC transformer, adding negligible runtime cost.

\begin{figure}[htbp]
  \centering
  \includegraphics[width=\linewidth]{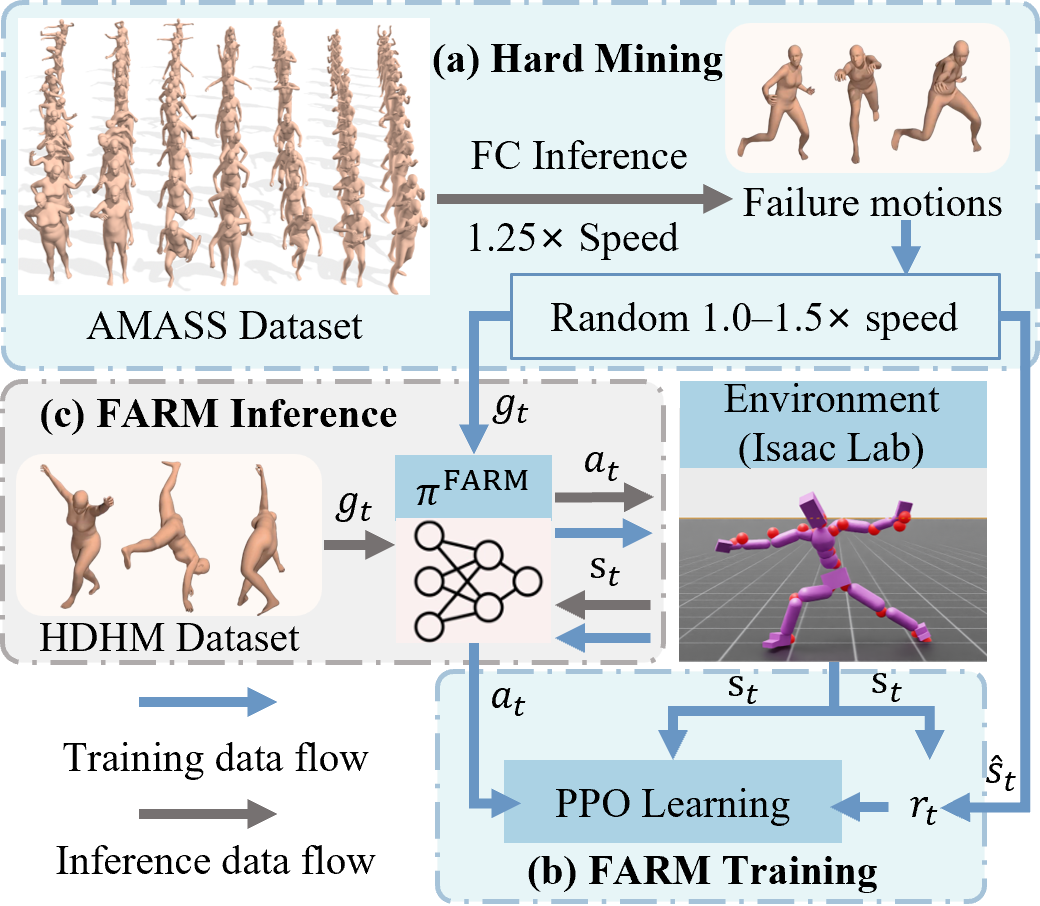}
  \caption{Overview of the FARM pipeline. Failure cases are mined by applying the FC model to the AMASS dataset at 1.25× speed. These hard samples are augmented with random 1.0–1.5× acceleration and used for FARM training. The learned policy is evaluated on the HDHM dataset to assess performance under high-dynamic motions.}
  % \caption{Overview of the proposed FARM framework. (a) Failure motions are identified by running the FC model on the AMASS dataset at 1.25× speed. These samples are then augmented via random 1.0–1.5× acceleration. (b) The FARM policy is trained in Isaac Lab using PPO. (c) The resulting controller is evaluated on the HDHM dataset to assess performance under high-dynamic motions.}
  \label{fig:total_framework}
\end{figure}

This section presents the details of our proposed FARM framework. Section \ref{subsec:framework} outlines the overall structure of FARM. Section \ref{subsec:FA} describes the frame-accelerated augmentation technique employed to expose the model to high-velocity pose transitions. Section \ref{subsec:Res-MoE} elaborates on the residual MoE architecture, which adaptively enhances the network's capacity to handle highly dynamic motions.

\subsection{Overall framework}
\label{subsec:framework}
As illustrated in Figure~\ref{fig:total_framework}, the training pipeline of the FARM framework first mines hard samples from AMASS, and then trains the FARM model on these samples. The trained policy is evaluated on HDHM to assess generalization to unseen high-dynamic motions.

Prior work \cite{luo2022universal, zhu2023neural} has shown that focusing on challenging samples can significantly improve model performance, and we observe similar benefits in our setting (see Section~\ref{subsec:ablation}). To mine such samples, we uniformly accelerate the entire AMASS dataset by a factor of 1.25× and apply the base FC \cite{tessler2024maskedmimic} controller $\pi^{FC}$ to perform inference on the accelerated motion sequences. Motions that the controller fails to track under this setting are collected as hard samples. A motion is considered a failure if the global mean per-joint position error exceeds 0.5 meters at any frame \cite{luo2021dynamics}. During training, they are augmented by randomly applying frame acceleration in the range of 1.0× to 1.5× (See Section \ref{subsec:FA}).
% These failure cases, often corresponding to rapid or complex transitions, serve as challenging training examples.

To train our controller for the motion imitation task, we adopt the framework of goal-conditioned reinforcement learning (GCRL). At each time step $t$, the agent receives the current observation $s_t$ along with a imitation goal signal $g_t$, and generates an action $a_t \sim \pi^{FARM}(a_t \mid s_t, g_t)$. After executing the action in the physics-based simulation environment (Issac Lab \cite{mittal2023orbit}), a transition to the next state $s_{t+1}$ occurs, and the agent receives a reward $r_t$. The objective is to optimize the policy $\pi^{FARM}$ to maximize the expected discounted return:
\begin{equation}
J = \mathbb{E}_{p}\left[ \sum_{t=0}^{T} \gamma^t r_t \right],
\end{equation}
where $p$ denotes the distribution over trajectories, and $\gamma \in [0, 1]$ is a discount factor. We use the proximal policy gradient (PPO \cite{schulman2017proximal}) to optimize $\pi^{FARM}$.

% We adopt the same definitions of state, action, and reward as used in MaskedMimic. 
%  where $f_t^{\text{char}}$ denotes the character’s body pose and velocity, and $f_t^{\text{scene}}$ corresponds to a local terrain heightmap.
%used to compute joint torques via a low-level controller
The state $s_t$ includes the character’s proprioception and terrain information, and is represented as
$
\left[ f_t^{\text{char}},\ f_t^{\text{scene}} \right]
$. Although evaluation is conducted on flat terrain, introducing irregular terrain during training improves the robustness of the policy when the reference motion contains slight ground penetration or floating artifacts. The goal $g_t$ encodes a sequence of future target poses from the reference motion.  The action $a_t$ represents the proportional-derivative (PD) target joint angles.
The reward $r_t$ encourages the agent to follow the reference motion while maintaining smooth and physically plausible behavior. It is defined as a weighted sum of a motion tracking term and an energy penalty term:
\begin{equation}
r_t = w^{\text{track}} r_t^{\text{track}}(s_t, \hat{s}_t) + w^{\text{energy}} r_t^{\text{energy}},
\end{equation}
where $r_t^{\text{track}}(s_t, \hat{s}_t)$ measures the discrepancy between the simulated character state $s_t$ and the reference state $\hat{s}_t$ \cite{da2008simulation, 10.1145/1833349.1781155, wang2020unicon}. The term $r_t^{\text{energy}}$ penalizes excessive joint torques to promote smoother motion. For detailed formulations of each component, refer to Section 5 of \citet{tessler2024maskedmimic}. Our implementation follows the same setting.

After training, the learned policy is evaluated on the HDHM dataset, which contains diverse high-dynamic motions that are not seen during training. Details of the HDHM dataset are provided in Section~\ref{subsec:datasets} and Appendix~A.

% Figure 2 illustrates the proposed FARM pipeline.  Each simulation step receives the same three input streams as FC—proprioception, a fifteen-frame reference window, and a \(16\times16\) terrain height grid—which are encoded separately and concatenated into a token sequence.  The sequence is simultaneously processed by the frozen FC transformer and by one speed-specialised expert transformer chosen by a lightweight gate.  The expert produces a residual embedding that is added element-wise to the FC embedding.  The shared decoder then maps the summed embedding to proportional–derivative (PD) joint targets.  Because the two transformers execute in parallel, wall-clock latency matches that of the original FC; the extra zero-initialised linear layer in each expert and the two-layer gate incur an overhead below \(0.1\) ms in our measurements.

\begin{figure*}[htbp]
  \centering
  \includegraphics[width=\linewidth]{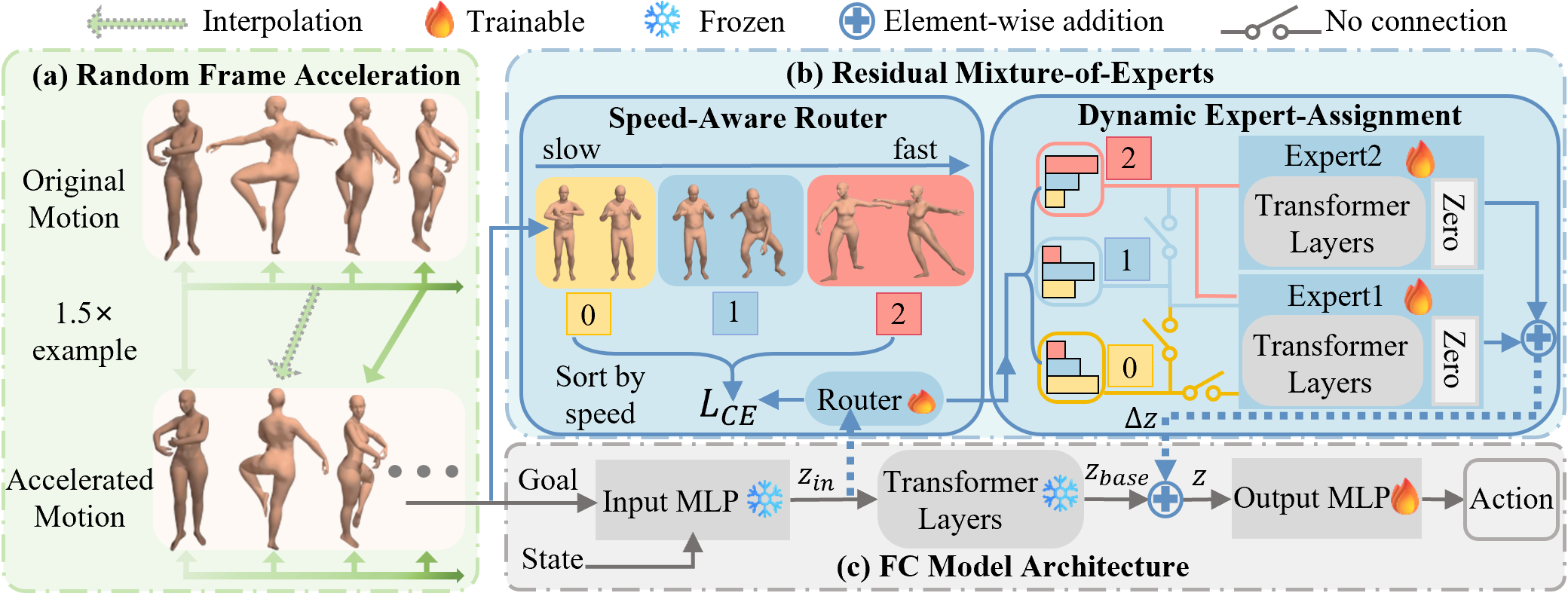}
  \caption{Overview of the FARM framework. Frame-accelerated augmentation increases frame intervals by uniformly downsampling motion sequences with random speed to simulate high-dynamic motions. The FARM model consists of a base FC model for general motion tracking and a residual MoE module that enhances high-dynamic motion control through the speed-aware router and dynamic expert-assignment.}
  \label{fig:framework}
\end{figure*}

\subsection{Frame-accelerated augmentation}
\label{subsec:FA}
For each motion clip used in training, a constant acceleration factor is drawn from a uniform distribution,
\(v\sim\mathcal{U}[1.0,1.5]\).
With the nominal playback step
\(\Delta t = \tfrac{1}{30}\,\mathrm{s}\) (30 Hz),
the resampled step becomes
\(\Delta t_{\text{new}} = v\,\Delta t\).
Virtual timestamps are defined as
\(t_k = k\,\Delta t_{\text{new}}\) for \(k\in\mathbb{N}\).
The pose \(\hat{f}_k\) at \(t_k\) is obtained by linear interpolation
between the two original frames
\(f_{\lfloor t_k/\Delta t \rfloor}\) and
\(f_{\lceil t_k/\Delta t \rceil}\).
The resulting sequence is fed to the controller at the original
30 Hz rate. Hence, consecutive poses correspond to a physical time
interval scaled by \(v\).
As illustrated in Figure \ref{fig:framework} (a), this augmentation
generates motions with enlarged inter‑frame displacements,
thereby exposing the policy to higher‑dynamic conditions.

% To expose the controller to high-dynamic motions, we introduce frame-accelerated augmentation during training. For each training sample, a speed factor $v$ is randomly sampled from a uniform distribution over the interval $[1.0,\ 1.5]$. Given a fixed playback framerate of 30 Hz, the original timestep between adjacent frames is $\Delta t = \frac{1}{30}$. After sampling $v$, the new timestep becomes $\Delta t_{\text{new}} = v \cdot \Delta t$. A new frame sequence is constructed by sampling frames from the original motion at virtual time points $t_k = k \cdot \Delta t_{\text{new}}$, where $k = 0, 1, 2, \dots$. Each sampled frame $\hat{f}_k$ is obtained via linear interpolation from the original motion sequence. Specifically, for a target time $t_k$, the two closest frames in the original motion, $f_{\lfloor t_k / \Delta t \rfloor}$ and $f_{\lceil t_k / \Delta t \rceil}$, are identified, and $\hat{f}_k$ is computed as their weighted average based on the relative distance to $t_k$. As shown in Figure \ref{fig:framework}(a), after resampling, the resulting motion is played back at the original framerate of 30 Hz, effectively increasing the temporal spacing between consecutive physical poses. This augmentation simulates the larger inter-frame differences characteristic of high-speed motion, thereby encouraging the policy to learn under high-dynamic conditions.

\subsection{Residual mixture-of-experts}
\label{subsec:Res-MoE}
The baseline FC controller tracks low‑dynamic motions reliably but fails on rapid pose changes.
To extend its dynamic range without disrupting its proven behaviour, we freeze the FC input MLP and Transformer \cite{vaswani2017attention}, update only the output MLP, and attach a residual MoE module (Figure \ref{fig:framework} (c)).
Given token features \(z_{\text{in}}\) from the frozen input MLP, the frozen Transformer produces
$z_{\text{base}}=\mathrm{Transformer}_{\mathrm{FC}}(z_{\text{in}})$.
In parallel the MoE computes a residual
\(\Delta z=\mathrm{MoE}(z_{\text{in}})\).
The combined representation
$
  z=z_{\text{base}}+\Delta z
$
is fed to the output MLP to yield the action \(a_t\).
Thus, the original controller is preserved, while the residual MoE supplies extra corrections for high‑dynamic segments. The residual MoE is equipped with a speed‑aware router and a dynamic expert‑assignment mechanism, described in the following paragraph.
% While the base FC controller demonstrates strong performance in tracking common low-dynamic motions, it exhibits limitations when dealing with rapid pose transitions in high-dynamic motions. To improve its high-dynamic tracking capacity while preserving its well-established tracking behavior, we adopt a fine-tuning strategy. As shown in Figure~\ref{fig:framework}(c), the input MLP and transformer layers of the FC model are kept frozen, and only the output MLP is updated during training. On top of this partially frozen architecture, we introduce a residual MoE module to enhance high-dynamic tracking performance.
% Let $z_{\text{in}}$ denote the input token features produced by the input MLP. The FC transformer encodes $z_{\text{in}}$ into a feature representation $z_{\text{base}} = \text{Transformer}_{\text{FC}}(z_{\text{in}})$. Simultaneously, the Residual MoE processes the same input and outputs a residual term $\Delta z = \text{MoE}(z_{\text{in}})$. The final representation is then computed as
% $
% z = z_{\text{base}} + \Delta z,
% $
% which is passed to the output MLP to generate the action $a_t$.
% This residual formulation enables the controller to retain the stable behavior of the FC model while selectively injecting expert-driven adjustments under high-dynamic conditions.

% The residual MoE module consists of a speed-aware router and a dynamic expert assignment mechanism. 
\paragraph{Speed-aware router (SAR)}
SAR promotes expert specialization by routing inputs according to motion intensity, allowing each expert to focus on a specific velocity regime. To guide this routing behavior, an auxiliary supervision signal is introduced. Specifically, for each training batch, the average joint velocity of the reference frame is computed for each sample. Samples are then sorted and evenly partitioned into three speed groups, labeled as 0, 1, and 2. These labels are used to supervise the router via a cross-entropy loss $\mathcal{L}_{\text{CE}}$.
The router is jointly optimized with the main reinforcement learning objective using the combined loss:
\begin{equation}
\mathcal{L}_{\text{router}} = \mathcal{L}_{\text{RL}} + \lambda_{\text{speed}} \mathcal{L}_{\text{CE}},
\end{equation}
where $\lambda_{\text{speed}}$ balances the auxiliary supervision. This formulation ensures that each expert receives data in a specific speed band, encouraging specialization and avoiding the common issue \cite{mu2025comprehensive} in unconstrained MoE setups where a single expert dominates and others are rarely selected.

\paragraph{Dynamic expert-assignment (DEA)}
DEA mechanism is inspired by human attention allocation: routine actions require minimal focus, whereas high-dynamic activities demand greater cognitive effort. Similarly, our method activates a variable number of experts based on motion intensity. 
Let $p \in \mathbb{R}^E$ be the router’s softmax output, where $E$ is the total number of experts. We define $k = \arg\max(p)$ and activate experts indexed from $1$ to $k$. The residual is computed as
\begin{equation}
\Delta z = \sum_{i=1}^{k} w_i \cdot \text{Expert}_i(z_{\text{in}}), \quad w_i = \sum_{j=i}^{E-1} p_j.
\end{equation}
When $k = 0$, no experts are used and the model reduces to the base FC controller. Followed by ControlNet \cite{zhang2023adding}, each expert copies the transformer architecture and weights of FC and appends a zero-initialized linear projection layer, ensuring the residual path does not affect the initial behavior. This design enables adaptive capacity scaling with motion difficulty while preserving the reliability of the original controller.

\section{Experiments}\label{sec:experiments}
\begin{figure}[htbp]
  \centering
  \includegraphics[width=\linewidth]{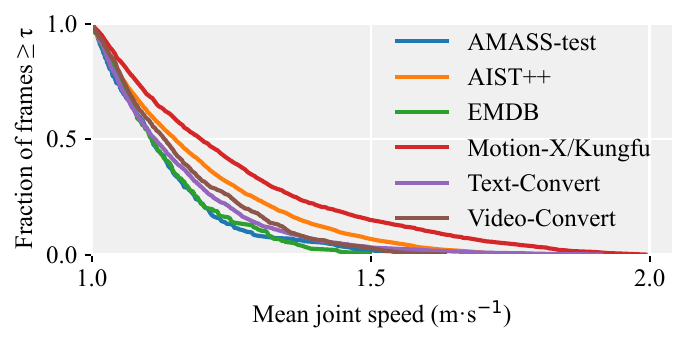}
  \caption{Survival functions of per-frame mean joint speeds for six datasets. $\tau$ represents the speed threshold. For visual clarity, only speeds between 1.0--2.0m$\cdot$s$^{-1}$ are shown. Almost all five HDHM-derived datasets lie above the AMASS-test curve, indicating greater dynamic activity.}
  \label{fig:speed_survival}
\end{figure}

% ======================================================
% Table 1  HDHM five subsets + Total (double-column)
% ======================================================

\begin{table*}[htbp]
  \centering
  \small
  \caption{
Comparison of our method with PHC+ and FC on the five subsets of the HDHM dataset and the combined overall set, with all results averaged over four random seeds.
\textuparrow indicates that higher values are better, while \textdownarrow indicates that lower values are better. 
Bold numbers highlight the best performance in each column. 
Our method consistently outperforms baselines across all subsets and the overall average. The results for PHC+ and FC are reproduced using the official released weights evaluated locally.
}
  \label{tab:hdhm_results}
  \begin{subtable}[t]{\linewidth}
    \centering
    \begin{tabular}{lccccccccc}
      \toprule
      & \multicolumn{3}{c}{AIST++} 
      & \multicolumn{3}{c}{EMDB} 
      & \multicolumn{3}{c}{Motion-X/Kungfu} \\
      \cmidrule(lr){2-4} \cmidrule(lr){5-7} \cmidrule(lr){8-10}
        & Success\,$\uparrow$  & MPJPE$_g$\,$\downarrow$ & MPJPE$_l$\,$\downarrow$
            & Success\,$\uparrow$  & MPJPE$_g$\,$\downarrow$ & MPJPE$_l$\,$\downarrow$
            & Success\,$\uparrow$  & MPJPE$_g$\,$\downarrow$ & MPJPE$_l$\,$\downarrow$ \\
      \midrule
      PHC\texttt{+} & 81.5\% & 210.7 & 102.0 
                   & 71.0\% & 302.8 &  89.5 
                   & 70.2\% & 265.5 & 122.4 \\
      FC            & 93.1\% & 108.5 &  76.6 
                   & 95.5\% &  98.3 &  57.4 
                   & 84.1\% & 155.3 &  86.5 \\
      % Ours          & 95.6\% &  94.2 &  69.6 
      %              & 97.7\% &  91.2 &  57.0 
      %              & 90.7\% & 124.2 &  72.0 \\
    Ours & \textbf{95.6\%} & \textbf{94.2} & \textbf{69.6} 
              & \textbf{97.7\%} & \textbf{91.2} & \textbf{57.0} 
              & \textbf{90.7\%} & \textbf{124.2} & \textbf{72.0} \\

      \bottomrule
    \end{tabular}
  \end{subtable}

  \vspace{0.5ex}

  \begin{subtable}[t]{\linewidth}
    \centering
    \begin{tabular}{lccccccccc}
      % \toprule
      & \multicolumn{3}{c}{Text-Convert} 
      & \multicolumn{3}{c}{Video-Convert} 
      & \multicolumn{3}{c}{Total(HDHM)} \\
      \cmidrule(lr){2-4} \cmidrule(lr){5-7} \cmidrule(lr){8-10}
        & Success\,$\uparrow$  & MPJPE$_g$\,$\downarrow$ & MPJPE$_l$\,$\downarrow$
            & Success\,$\uparrow$  & MPJPE$_g$\,$\downarrow$ & MPJPE$_l$\,$\downarrow$
            & Success\,$\uparrow$  & MPJPE$_g$\,$\downarrow$ & MPJPE$_l$\,$\downarrow$ \\
      \midrule
      PHC\texttt{+} & 87.3\% & 166.0 &  77.3 
                   & 57.8\% & 303.4 & 144.5 
                   & 80.3\% & 217.6 & 100.6 \\
      FC            & 96.6\% &  76.9 &  50.7 
                   & 83.2\% & 173.0 &  91.4 
                   & 92.3\% & 111.3 &  71.7 \\
      % Ours          & 98.6\% &  71.2 &  48.5 
      %              & 90.7\% & 116.8 &  71.8 
      %              & 95.6\% &  95.0 &  64.4 \\
      Ours & \textbf{98.6\%} & \textbf{71.2} & \textbf{48.5} 
              & \textbf{90.7\%} & \textbf{116.8} & \textbf{71.8} 
              & \textbf{95.6\%} & \textbf{95.0} & \textbf{64.4} \\

      \bottomrule
    \end{tabular}
  \end{subtable}
\end{table*}

\begin{figure*}[!h]
  \centering
  \includegraphics[width=\linewidth]{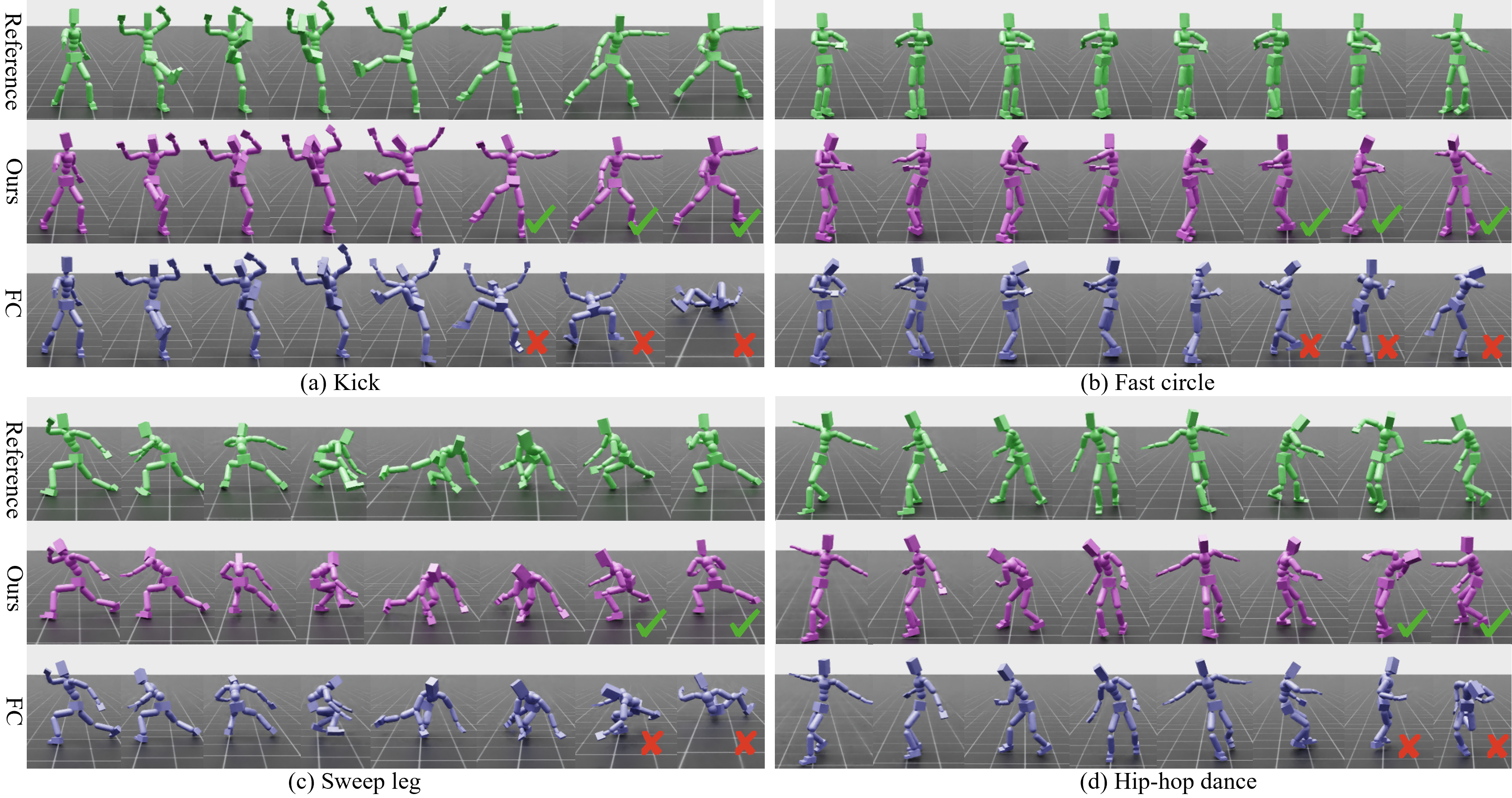}
  \caption{Qualitative comparison on four high-dynamic motion clips. Motion frames are presented in playback order from left to right. The reference motion is shown in green, our FARM results in purple, and the FC baseline in blue. FARM consistently maintains accurate motion tracking (green check), while FC exhibits noticeable failures under rapid pose changes (red cross).}
  \label{fig:qual_frames}
\end{figure*}

In this section, we conduct a comprehensive evaluation of the proposed FARM framework. Section~4.1 describes the datasets used in our experiments. Section~4.2 presents both quantitative and qualitative comparisons. Section~4.3 analyzes the expert selection behavior within the residual MoE module. Section~4.4 provides ablation studies to investigate the contribution of each component in the FARM framework. Additional experiment details are provided in Appendix B.

\subsection{Datasets}
\label{subsec:datasets}
Following \citet{luo2024universal}, we filter the AMASS dataset to remove physically implausible motions, and split it into two subsets: AMASS-train for training and AMASS-test for evaluation. 
To further assess high-dynamic motion tracking, we construct the HDHM dataset, composed of curated clips from five sources: AIST++ \cite{li2021learn, tsuchida2019aist}, EMDB \cite{kaufmann2023emdb}, Motion-X Kungfu subset \cite{lin2023motion}, Text-Convert, and Video-Convert. All clips are manually filtered to exclude those involving explicit environment interaction or exhibiting non-physical artifacts such as body self-intersections, floating, or ground penetration. All sequences are temporally resampled to 30 Hz.
AIST++ contains diverse dance sequences, EMDB covers both indoor daily activities and outdoor sports, and the Motion-X Kungfu subset features complex kungfu movements. The Text-Convert set is generated using motion diffusion models \cite{tevet2022human, tevet2024closdclosingloopsimulation} conditioned on high-dynamic prompts, while the Video-Convert set is created by converting YouTube martial arts videos via the GVHMR model \cite{shen2024gvhmr}. 
Figure~\ref{fig:speed_survival} illustrates that the HDHM dataset exhibits higher motion dynamics compared to AMASS.
Among these, AIST++, EMDB, and the Motion-X Kungfu subset are publicly available datasets, while Text-Convert and Video-Convert are constructed by us.
In total, the HDHM dataset contains 3593 clips with an average duration of 9.4 seconds, with AIST++ (1320, 12.4s), EMDB (45, 36.3s), Motion-X Kungfu (663, 10.1s), Text-Convert (1392, 5.9s), and Video-Convert (173, 5.5s). For more details, refer to Appendix A.

\subsection{Performance Evaluation}

\begin{figure*}[!h]
  \centering
  \includegraphics[width=\linewidth]{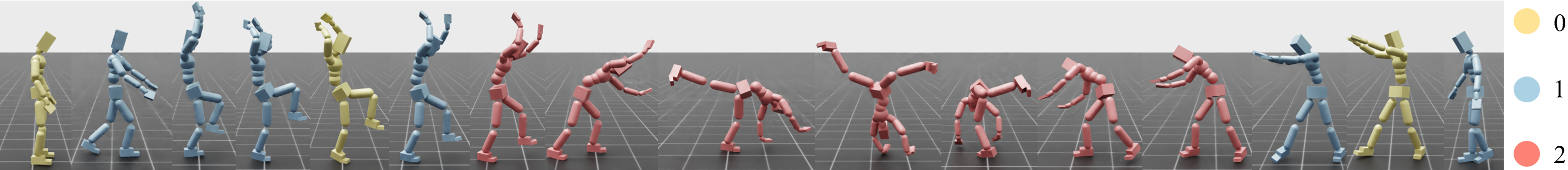}
  \caption{Visualization of expert activation throughout a cartwheel motion. 
Each frame is colored according to the number of active experts (yellow for 0, blue for 1, and red for 2). 
We observe that expert activation aligns with motion dynamics.
}% used by the residual MoE
  \label{fig:case_gate}
\end{figure*}

\begin{table}[htbp]
  \centering
  \small
  \caption{
  Performance on AMASS‐train and AMASS‐test datasets. 
  The AMASS dataset is primarily composed of low-dynamic motions. 
  Our method achieves high tracking success rates comparable to the baselines, 
  and the increase in MPJPE$_g$ compared to FC remains below 2\,mm.
  }
  \label{tab:amass_train_test}
  \begin{tabular}{lcccc}
    \toprule
    & \multicolumn{2}{c}{AMASS‐train} & \multicolumn{2}{c}{AMASS‐test} \\
    \cmidrule(lr){2-3}\cmidrule(lr){4-5}
     & Success\,$\uparrow$ & MPJPE$_g$\,$\downarrow$  & Success\,$\uparrow$  & MPJPE$_g$\,$\downarrow$  \\
    \midrule
    PHC+ & \textbf{99.9\%} & \textbf{28.8} & 98.5\% & \textbf{38.1} \\
    FC   & 99.7\% & 33.3 & \textbf{100\%} & 40.8 \\
    Ours & \textbf{99.9\%} & 35.2 & \textbf{100\%} & 42.1 \\
    \bottomrule
  \end{tabular}
\end{table}

\paragraph{Quantitative comparison.}
We compare FARM with two state-of-the-art baselines, FC \cite{tessler2024maskedmimic} and PHC$+$ \cite{luo2024universal}, on both the HDHM and AMASS datasets. 
Evaluation metrics include the motion tracking success rate, the global mean per-joint position error (MPJPE$_g$), and the root-relative MPJPE (MPJPE$_\ell$), following the definitions used in \citet{luo2021dynamics}. 
The success rate measures whether the humanoid can track the reference motion without falling or significantly lagging behind. 
MPJPE$_\ell$ and MPJPE$_g$ quantify the imitation accuracy in root-relative and global coordinates, respectively. 
Table~\ref{tab:hdhm_results} summarizes the performance on the five subsets of the HDHM dataset and the overall aggregate. 
Our method achieves the highest success rate and lowest MPJPE scores on all subsets, demonstrating improved tracking robustness and accuracy, especially under high-dynamic motions.Table~\ref{tab:amass_train_test} shows that our method retains high success rates comparable to the baselines, and introduces less than a 2mm increase in MPJPE$_g$ relative to FC.
This shows that FARM not only improves tracking on challenging motions but also preserves performance on standard motions.

\paragraph{Qualitative comparison.} 
To further assess high-dynamic tracking quality, we visually compare FARM with the FC baseline on four challenging clips in Figure~\ref{fig:qual_frames}. While FC often drifts or collapses during fast transitions, FARM successfully maintains balance and closely follows the reference, demonstrating improved robustness under rapid pose changes.

% ======================================================
% Table 2  AMASS train / test (single-column)
% ======================================================

\subsection{Expert behaviour analysis}
\paragraph{Case study.}
We present a case study on a cartwheel motion to illustrate how the residual MoE dynamically adjusts expert usage. As shown in Figure~\ref{fig:case_gate}, each humanoid is colored based on the number of activated experts. We observe that expert activations correlate well with motion dynamics. Fewer experts are used during low-motion phases, while high-dynamic segments trigger more experts. This indicates that the SAR successfully learns to distinguish motion dynamics, providing a foundation for effective dynamic expert-assignment.

\begin{table*}[htbp]
  \centering
  \small
  \setlength{\tabcolsep}{4pt} % 适当收窄列间距
  \begin{tabular}{c c c c c c c c c | cc | cc}
    \toprule
     &
    \multicolumn{1}{c}{Data} &
    \multicolumn{3}{c}{MoE} &
    \multicolumn{4}{c|}{Router} &
    \multicolumn{2}{c|}{HDHM} &
    \multicolumn{2}{c}{AMASS-test} \\
    \cmidrule(lr){2-2}
    \cmidrule(lr){3-5}
    \cmidrule(lr){6-9}
    \cmidrule(lr){10-11}
    \cmidrule(lr){12-13}
   Row  &
    FA & Res-MoE & Full-MoE & No-MoE
       & SAR & DEA & Top1 & Top2
       & Success\,$\uparrow$ & MPJPE\textsubscript{g}\,$\downarrow$
       & Success\,$\uparrow$ & MPJPE\textsubscript{g}\,$\downarrow$ \\
    \midrule
    % ------------------- Baseline -------------------
    1 & \xmark & \xmark & \xmark & \xmark & \xmark & \xmark & \xmark & \xmark &
      92.3\% & 111.3 & 100\% & 40.8 \\
    \midrule
    % ------------------- Variants -------------------
    2 & \xmark & \cmark & \xmark & \xmark & \cmark & \cmark & \xmark & \xmark &
      95.1\%\,\textcolor{blue}{(+3.0\%)} &
       97.0\,\textcolor{blue}{($-$12.9\%)} &
      100\% & 41.6\,\textcolor{red}{(+2.0\%)} \\
    3 & \cmark & \xmark & \cmark & \xmark & \cmark & \cmark & \xmark & \xmark &
      94.3\%\,\textcolor{blue}{(+2.2\%)} &
      104.9\,\textcolor{blue}{($-$5.8\%)} &
      100\% & 47.3\,\textcolor{red}{(+15.9\%)} \\
    4 & \cmark & \cmark & \xmark & \xmark & \xmark & \cmark & \xmark & \xmark &
      94.8\%\,\textcolor{blue}{(+2.7\%)} &
       97.5\,\textcolor{blue}{($-$12.4\%)} &
      100\% & 44.6\,\textcolor{red}{(+9.3\%)} \\
    5 & \cmark & \xmark & \xmark & \cmark & \xmark & \xmark & \xmark & \xmark &
      92.8\%\,\textcolor{blue}{(+0.5\%)} &
      111.4\,\textcolor{red}{(+0.1\%)} &
      100\% & 46.2\,\textcolor{red}{(+13.2\%)} \\
    6 & \cmark & \cmark & \xmark & \xmark & \cmark & \xmark & \cmark & \xmark &
      95.5\%\,\textcolor{blue}{(+3.5\%)} &
       97.5\,\textcolor{blue}{($-$12.4\%)} &
      100\% & 44.2\,\textcolor{red}{(+8.3\%)} \\
    7 & \cmark & \cmark & \xmark & \xmark & \cmark & \xmark & \xmark & \cmark &
      95.3\%\,\textcolor{blue}{(+3.3\%)} &
       98.1\,\textcolor{blue}{($-$11.9\%)} &
      100\% & 44.8\,\textcolor{red}{(+9.8\%)} \\
    8 & \cmark & \cmark & \xmark & \xmark & \cmark & \cmark & \xmark & \xmark &
      95.6\%\,\textcolor{blue}{(+3.6\%)} &
       95.0\,\textcolor{blue}{($-$14.6\%)} &
      100\% & 42.1\,\textcolor{red}{(+3.2\%)} \\
    \bottomrule
\end{tabular}
  \caption{Ablation results on the HDHM and AMASS-test datasets. 
Row 1 shows the performance of the baseline FC model. 
Values in parentheses indicate relative change with respect to this baseline. 
Blue text denotes improvement, while red text indicates degradation. 
We ablate the effects of frame acceleration (FA), different MoE designs—including No MoE (no expert module), Res-MoE (our residual formulation), Full-MoE (jointly trained base and expert networks), and router configurations including speed-aware router (SAR), dynamic expert-assignment (DEA), and hard expert selection strategies (Top1 and Top2).}
  \label{tab:ablation}
\end{table*}

\begin{figure}[!tb]
  \centering
  \includegraphics[width=\linewidth]{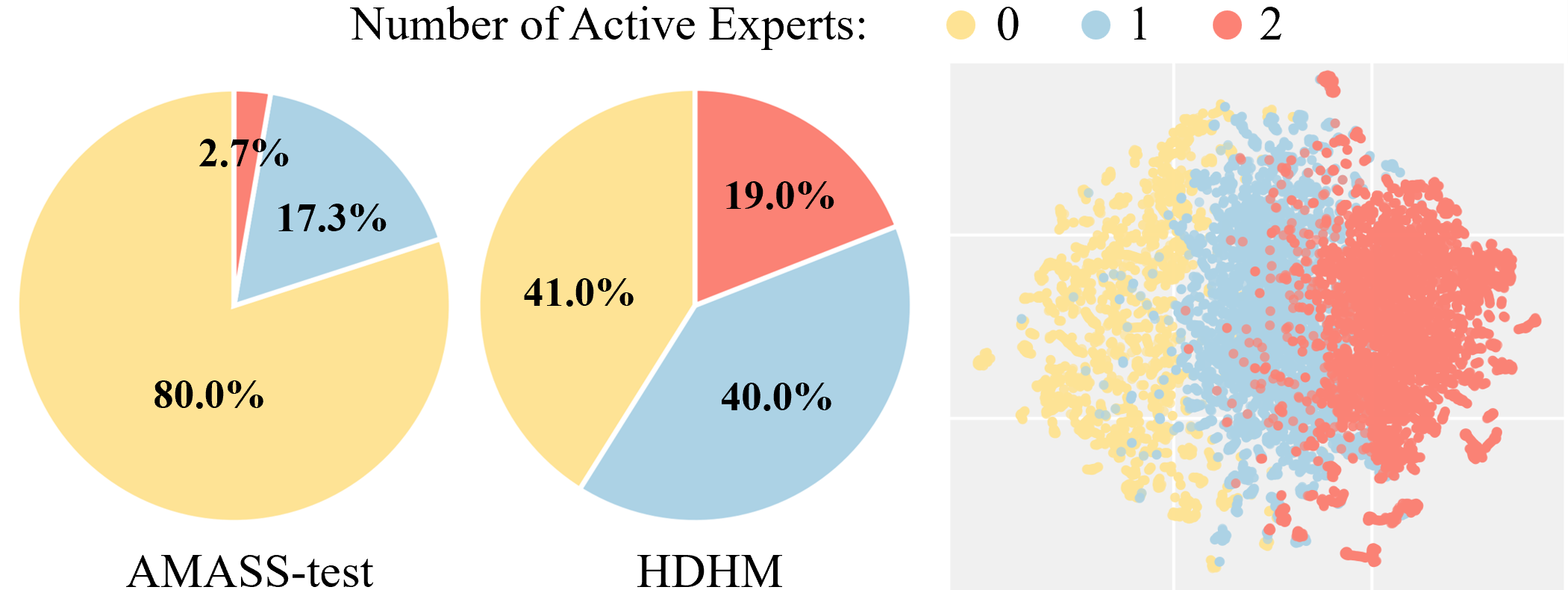}
\caption{\textbf{Left:} Distribution of expert usage indicates that our controller engages more experts on HDHM in response to higher motion dynamics. \textbf{Right:} t-SNE visualization of Video-Convert embeddings demonstrates that the residual MoE encodes distinct and structured features.}
  \label{fig:expert_pie_tsne}
\end{figure}
\paragraph{Statistical and spatial analysis.}
To further understand the behavior of our residual MoE, we analyze both the distribution of activated expert counts and the spatial organization of expert outputs. As shown in Figure~\ref{fig:expert_pie_tsne} (left), the AMASS-test set, which primarily contains low-dynamic motions, rarely activates experts, with 80\% of frames using only the base controller. In contrast, the HDHM dataset exhibits more frequent activation of experts, with over half of the frames involving one or two experts. This confirms that our model allocates more expert capacity for high-dynamic motions.

To investigate whether these expert activations lead to distinguishable control behaviors, we extract the output embeddings of the residual MoE under different expert counts, and visualize them using t-SNE, as shown in Figure~\ref{fig:expert_pie_tsne} (right). Each point represents the final embedding $z$ from a motion frame, colored by the number of activated experts. The resulting projection reveals a clear spatial separation, indicating that the residual MoE learns structured and distinct representations corresponding to different motion complexities. This further supports the effectiveness of our SAR and DEA mechanism.

\subsection{Ablation study}
\label{subsec:ablation}

\paragraph{Component ablation.}
We group the ablation results in Table~\ref{tab:ablation} by major components. Row~5 shows that applying frame acceleration (FA) alone yields minimal improvement, indicating that the base model cannot fully exploit high-dynamic data. In contrast, combining FA with MoE (Rows~3 and 8) leads to clear gains, suggesting that additional expert capacity is essential for leveraging harder samples.
Among MoE designs, the residual variant (Res-MoE, Rows~2 and 8) consistently outperforms both the No-MoE baseline (Row~5) and the Full-MoE variant (Row~3), which jointly updates the base and expert modules. Full-MoE degrades performance on AMASS-test, highlighting that modifying the pretrained base harms generalization. In contrast, Res-MoE preserves the base and adapts only the residual, improving high-dynamic performance without sacrificing low-dynamic accuracy.
Router-wise, SAR proves crucial, as its absence (Row~4) leads to notable drops. DEA (Row~8) surpasses both Top1 (Row~6), which overloads a single expert, and Top2 (Row~7), which involves unnecessary experts. DEA better balances expert capacity based on motion intensity.

\begin{figure}[htbp]
  \centering
  \includegraphics[width=\linewidth]{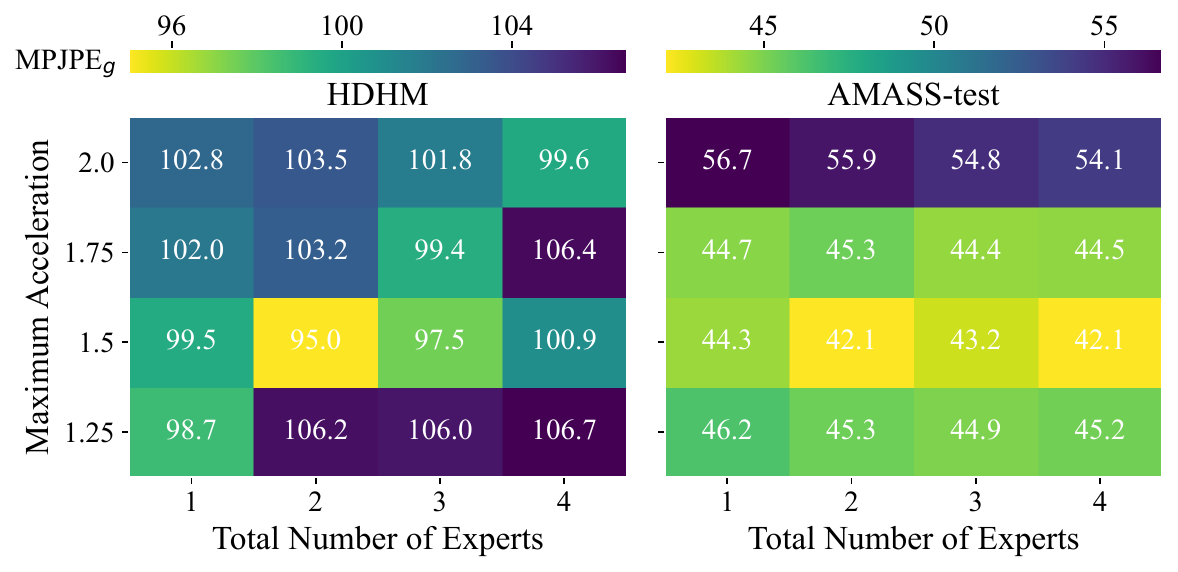}
  \caption{Heatmaps of MPJPE$_g$ across total number of experts and maximum acceleration factor for the HDHM and AMASS‐test datasets.}
  \label{fig:mpjpe_heatmaps}
\end{figure}

\paragraph{Impact of frame acceleration and number of experts.}
Figure~\ref{fig:mpjpe_heatmaps} presents heatmaps analyzing the impact of two key design choices: the maximum random frame acceleration factor and the total number of experts in the residual MoE. The results show a clear trend that both insufficient or excessive acceleration, as well as too few or too many experts, can lead to suboptimal performance.
On the one hand, a maximum acceleration factor that is too low fails to introduce sufficient high-dynamic diversity, while excessively large factors may generate physically implausible motions that hinder learning. On the other hand, using only one expert overloads it with a wide range of dynamics, while having too many experts results in over-fragmented expert assignments, making it harder for each expert to learn meaningful patterns.

\begin{table}[htbp]
  \centering
  \small  % 缩小表格字体
  \setlength{\tabcolsep}{2pt} % 适当收窄列间距
  \caption{Comparison of fine-tuning on full data versus failure data. Fine-tuning on failure data achieves slightly higher success rates, lower MPJPE$_g$, and dramatically faster convergence, demonstrating the efficiency of targeted fine-tuning on challenging samples. All experiments are conducted on a single RTX 4090 GPU.}
  % \caption{Comparison of fine-tuning on full data versus failure data. Fine-tuning on failure data achieves slightly higher success rates, lower MPJPE$_g$, and dramatically faster convergence, demonstrating the efficiency of targeted fine-tuning on challenging samples.}
  \label{tab:fine_tune_comparison}
  \begin{tabular}{lccccc}
    \toprule
          & \multicolumn{2}{c}{HDHM} & \multicolumn{2}{c}{AMASS-test} & AMASS-train  \\
    \cmidrule(lr){2-3}\cmidrule(lr){4-5}\cmidrule(lr){6-6}
    Data  & Success\,$\uparrow$ & MPJPE\textsubscript{g}\,$\downarrow$ & Success\,$\uparrow$ & MPJPE\textsubscript{g}\,$\downarrow$ & Training time \\
    \midrule
    Full    & 95.1\% & 99.2 & \textbf{100\%} & 44.4 & 35\,h \\
    Failure & \textbf{95.6\%} & \textbf{95.0} & \textbf{100\%} & \textbf{42.1} & \textbf{6\,h} \\
    \bottomrule
\end{tabular}
\end{table}

\paragraph{Fine-tuning data.}
Table~\ref{tab:fine_tune_comparison} compares full-data fine-tuning with fine-tuning only on failure cases. Fine-tuning on failure data achieves slightly better performance on both the HDHM and AMASS-test datasets, with higher success rates and lower MPJPE$_g$. Notably, this targeted approach also reduces the training time from 35 hours to just 6 hours, representing a substantial gain in efficiency. These results suggest that difficult samples provide more informative gradients and that focusing training on such challenging segments can accelerate convergence while improving final performance.

\section{Conclusion}\label{sec:conclusion}
We present FARM, a high-dynamic humanoid control framework that combines frame-accelerated augmentation with a residual MoE architecture. To evaluate high-dynamic performance, we curate the HDHM dataset, which contains diverse, high-velocity motion clips collected from multiple sources. Experimental results demonstrate that FARM significantly improves tracking accuracy on HDHM while maintaining strong generalization on AMASS.
Despite these advances, our framework still exhibits failure cases on HDHM, particularly on motions with subtle artifacts such as ground penetration, floating, or joint jitter. While these artifacts may appear visually acceptable, they pose non-trivial challenges for controllers. Future work will investigate how to eliminate such artifacts at the data generation stage and explore architectural improvements to enhance robustness against noise. Another important direction is to deploy the FARM framework on real-world humanoid robots.

\bibliography{aaai2026}

\clearpage
\appendix
\section{HDHM Datasets}

The HDHM dataset comprises 3593 motion clips sourced from five datasets: AIST++~\cite{li2021learn, tsuchida2019aist}, EMDB~\cite{kaufmann2023emdb}, Motion-X Kungfu~\cite{lin2023motion}, and two in-house collections, Text-Convert and Video-Convert. All clips are temporally resampled to 30 Hz and stored in a unified SMPL format \cite{10.1145/2816795.2818013}.

\begin{figure}[htbp]
  \centering
  \includegraphics[width=\linewidth]{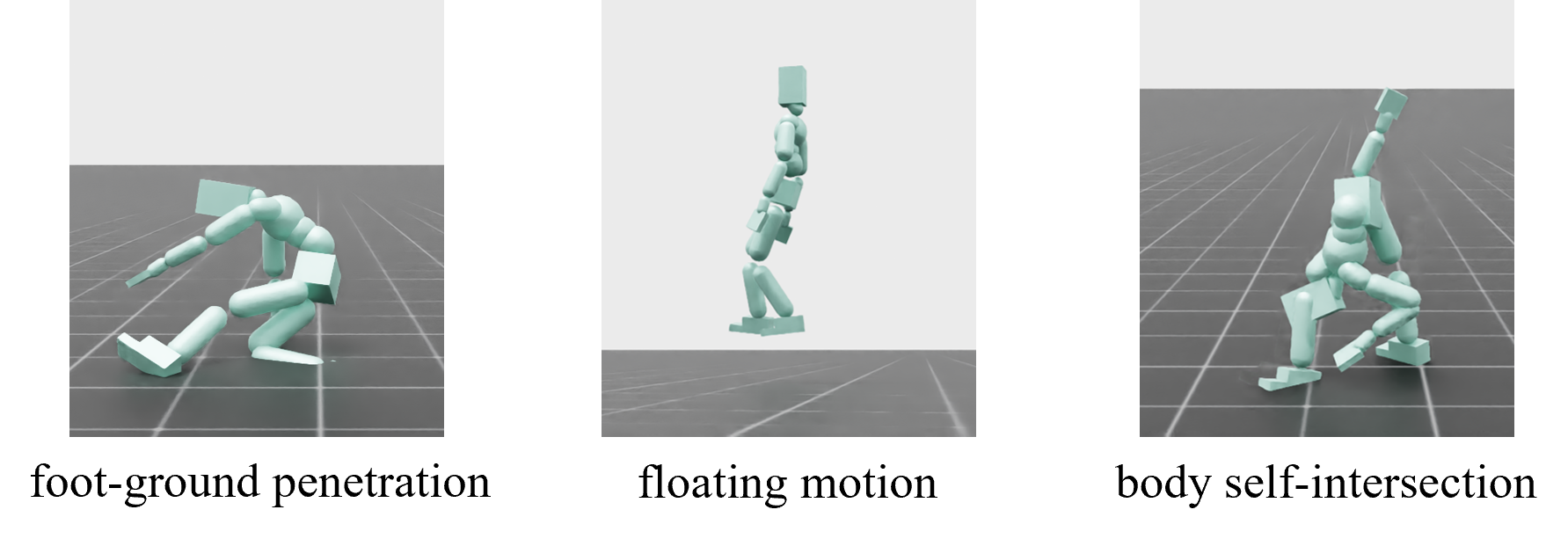} 
  \caption{Examples of filtered clips with non-physical artifacts. Left: foot-ground penetration. Middle: floating motion. Right: body self-intersection.}
  \label{fig:nonphysical_examples}
\end{figure}

To ensure physical plausibility while maintaining motion diversity, we first run the FC baseline policy on all candidate sequences and collect those with significant tracking failures. We then manually inspect these failure cases and remove clips that exhibit visually obvious non-physical artifacts. Specifically, we exclude motions with ground penetration, floating, severe joint jitter, or self-intersections (Figure \ref{fig:nonphysical_examples}), while tolerating mild imperfections that are not perceptually distracting.
The following criteria guide our manual filtering process:
\begin{itemize}
    \item \textbf{No ground penetration}: feet and body parts must remain above the ground plane.
    \item \textbf{No floating}: characters should not hover or lose ground contact unnaturally.
    \item \textbf{No body self-intersection}: limbs should not pass through each other.
\end{itemize}

Figure~\ref{fig:nonphysical_examples} shows examples of such non-physical artifacts removed during curation.

For publicly available datasets (AIST++, EMDB, and Motion-X Kungfu), we provide an ignore list of file names to indicate the filtered sequences, allowing reproducibility when using the full released dataset. For the synthetic sets (Text-Convert and Video-Convert), we directly release the fully cleaned versions.

\paragraph{Text-Convert.} 
This subset is generated using motion diffusion models~\cite{tevet2022human, tevet2024closdclosingloopsimulation} conditioned on text prompts describing high-dynamic actions (e.g., "The person swings head very quick to find something"). 
To construct such prompts, we leverage a large language model (Gemini) to generate diverse and expressive descriptions tailored to highly dynamic behaviors. 
These prompts are then fed into the diffusion model to synthesize motion sequences. 
For each prompt, we sample multiple generations and manually retain only the ones that satisfy physical plausibility and demonstrate sufficient dynamicity.

\paragraph{Video-Convert.} This subset is constructed by extracting martial arts motions from public YouTube videos and PKU-DyMVHumans videos \cite{zheng2024pku}. We apply the GVHMR model~\cite{shen2024gvhmr} to estimate SMPL motion from monocular RGB input. Due to noise in the video source and inaccuracies in model estimation, we retain only a small subset of high-quality clips that meet the aforementioned physicality criteria.

\section{Experimental Details}
\subsection{Platform Configuration}
All experiments are conducted on a workstation running Ubuntu 22.04, equipped with a 13th Gen Intel(R) Core(TM) i9-13900K CPU, a single NVIDIA RTX 4090 GPU, and 128 GB of RAM. The simulation environments are built using Isaac Lab, and the codebase is implemented in Python using PyTorch 2.5.1.

\subsection{Network Architecture}
Our policy network follows a transformer-based architecture augmented with a residual mixture-of-experts (MoE) module. The overall structure consists of a frozen base transformer and a set of learnable expert branches that operate in residual mode. The base transformer comprises 4 layers, each with 4 attention heads, a hidden size of 512, and feedforward layers of size 1024. An MLP first processes observation inputs with two hidden layers of size 256 and ReLU activations before being tokenized for the transformer. The output features are decoded through a three-layer MLP with hidden sizes of 1024, each followed by a ReLU activation, to generate the final actions.

The residual MoE module enhances the base transformer by dynamically injecting expert capacity depending on motion dynamics. It maintains 2 parallel transformer adapters, each followed by a zero-initialized projection layer. A router network takes the concatenated current and reference pose embeddings as input and outputs soft weights over 2 branches via a two-layer MLP with a hidden size of 256 and ReLU activation.

\subsection{Training Hyperparameters}
The training parameters are shown in Figure \ref{tab:training_hyperparams}.
\begin{table}[h]
\centering
\caption{Training hyperparameters used in our experiments.}
\label{tab:training_hyperparams}
\begin{tabular}{ll}
\toprule
\textbf{Hyperparameter}          & \textbf{Value}         \\
\midrule
Actor optimizer                 & Adam \cite{kingma2014adam}                   \\
Actor learning rate            & $1 \times 10^{-5}$     \\
Critic optimizer               & Adam                   \\
Critic learning rate           & $5 \times 10^{-5}$     \\
Discount factor $\gamma$       & 0.99                   \\
GAE parameter $\tau$           & 0.95                   \\
Clipping parameter $\epsilon$ & 0.2                    \\
Speed loss coefficient $\lambda_{speed}$         & 1.0    \\
Batch size                     & 2048                   \\
Number of environments         & 512                    \\
\bottomrule
\end{tabular}
\end{table}

% \bibliography{aaai2026}
% \end{document}

\end{document}